%% file: main.tex
\pdfoutput=1

\documentclass[11pt]{article}

\usepackage{emnlp2023}

\usepackage{times}
\usepackage{latexsym}
\usepackage{algorithm}
\usepackage{algpseudocode}
\usepackage{booktabs}
\usepackage{multirow}
\usepackage{amsmath}
\usepackage{amssymb}
\usepackage{xcolor}
\usepackage{url}
\usepackage{graphicx}
\usepackage[T1]{fontenc}
\usepackage[utf8]{inputenc}
\usepackage{microtype}
\usepackage{inconsolata}

\title{Indexing the Unreadable: LLM-Native Recursive Construction and Search of Service Taxonomies}

\author{
  {\bf Wei Zheng \quad Yang Yan \quad Yiyang Shao \quad Jinyang Li \quad Zeze Chang} \\
  {\bf Yukuang Jia \quad Qiming Mao \quad Chihyung Wang \quad Jingbin Zhou} \\[4pt]
  openJiuwen A2X \\[2pt]
  \texttt{zhengw@vt.edu}
}

\begin{document}
\maketitle

\input{paper_body}

\end{document}

%% file: paper_body.tex

\begin{abstract}
The era of the \emph{Internet of Agents (IoA)} is taking shape: LLM agents are expected to fulfill user goals by orchestrating fast-growing populations of Model Context Protocol (MCP) servers, Agent-to-Agent (A2A) endpoints, reusable skills, and other LLM-callable services. Yet LLMs face a structural mismatch with this regime: \emph{effective context is a scarce resource that does not scale with the number of services.} Concatenating thousands of service descriptions into a prompt overflows the context window, and even when the window is large enough, models systematically under-attend to information in the middle of long inputs, the well-documented Lost-in-the-Middle phenomenon. This is fundamentally a question of \textbf{context management for service discovery}. To address this, we propose an LLM-native progressive-disclosure scheme and its concrete instantiation, \textbf{A2X (Agent-to-Anything service discovery)}: an LLM-driven pipeline that automatically organizes the registered services into a hierarchical taxonomy and walks it layer by layer at query time, so that every LLM call sees only a small candidate set highly relevant to the user query. This decouples effective-context scarcity from registry size and significantly reduces token consumption while improving retrieval accuracy. Compared to full-context dumping, A2X achieves a 6.2-point Hit Rate gain at one-ninth the prompt-token cost; compared to the state-of-the-art open-source embedding-based baseline, A2X improves Hit Rate by more than 20 points.
\end{abstract}

\section{Introduction}
\label{sec:intro}

We are entering the era of the \emph{Internet of Agents (IoA)}, in which LLM agents are expected to take a goal stated in natural language and orchestrate the online services needed to fulfill it. The pool of such services is expanding rapidly. Model Context Protocol (MCP) servers \citep{anthropic2024mcp} alone now number in the thousands; Agent-to-Agent (A2A) endpoints \citep{google2025a2a}, function-tool catalogs, reusable skill packages \citep{anthropic2025skills}, and an increasing variety of other services are all becoming agent-callable.

However, this expansion collides directly with an intrinsic limitation of today's LLMs: \emph{effective context is a scarce resource that does not scale with the number of services.} Two failure modes follow. First, \emph{token bloat}: for an agent with 1{,}000 MCP tools, the catalog alone persistently occupies \textbf{about 500 K tokens} of context \citep{anthropic2025mcpcode,hasan2026mcpsmell}. Second, \emph{Lost-in-the-Middle} \citep{liu2024lostmiddle}: even when the window is nominally large enough, models systematically under-attend to information placed in the middle of long inputs, and recent long-context evaluations \citep{hsieh2024ruler} confirm that accuracy does not scale linearly with context length, so scarce \emph{effective} context, not raw window size, is the binding constraint.

Confronted with this mismatch, one line of work bypasses the LLM's context constraint via embedding-based retrieval \citep{gan2025ragmcp, fei2025mcpzero}, at the cost of giving up the LLM's semantic-understanding strength on cross-vocabulary, multi-constraint, and long-tail queries, which becomes a recall bottleneck in complex scenarios \citep{liu2025livemcpbench, shi2025toolret}. This paper takes the other path: building an efficient and accurate \textbf{LLM-native service discovery}, whose core research question is \textbf{context management for service discovery}: how should a large, dynamic registry be organized and selectively disclosed to an LLM so that every decision is taken over a short candidate set highly relevant to the user query, while the registry as a whole remains discoverable?

Recent work on tree-structured LLM reasoning, including Chain-of-Thought \citep{wei2022cot} and Tree-of-Thoughts \citep{yao2023tot}, has shown that decomposing a complete decision into a sequence of local decisions significantly improves LLM accuracy and controllability on complex reasoning tasks. We approach service discovery in the same spirit, as a problem to be solved by LLM-native progressive disclosure: organize the registry into a hierarchical \textbf{taxonomy} such that each local decision sees only one node's children rather than the whole structure; then, at query time, recursively descend the taxonomy along branches matching the query, so each LLM call weighs only a small candidate set highly relevant to the query. Service catalogs admit such a taxonomy naturally: Travel divides into flights, hotels, and currency; Finance into payments, lending, and analytics.

\begin{figure*}[t]
\centering
\includegraphics[width=0.8\textwidth]{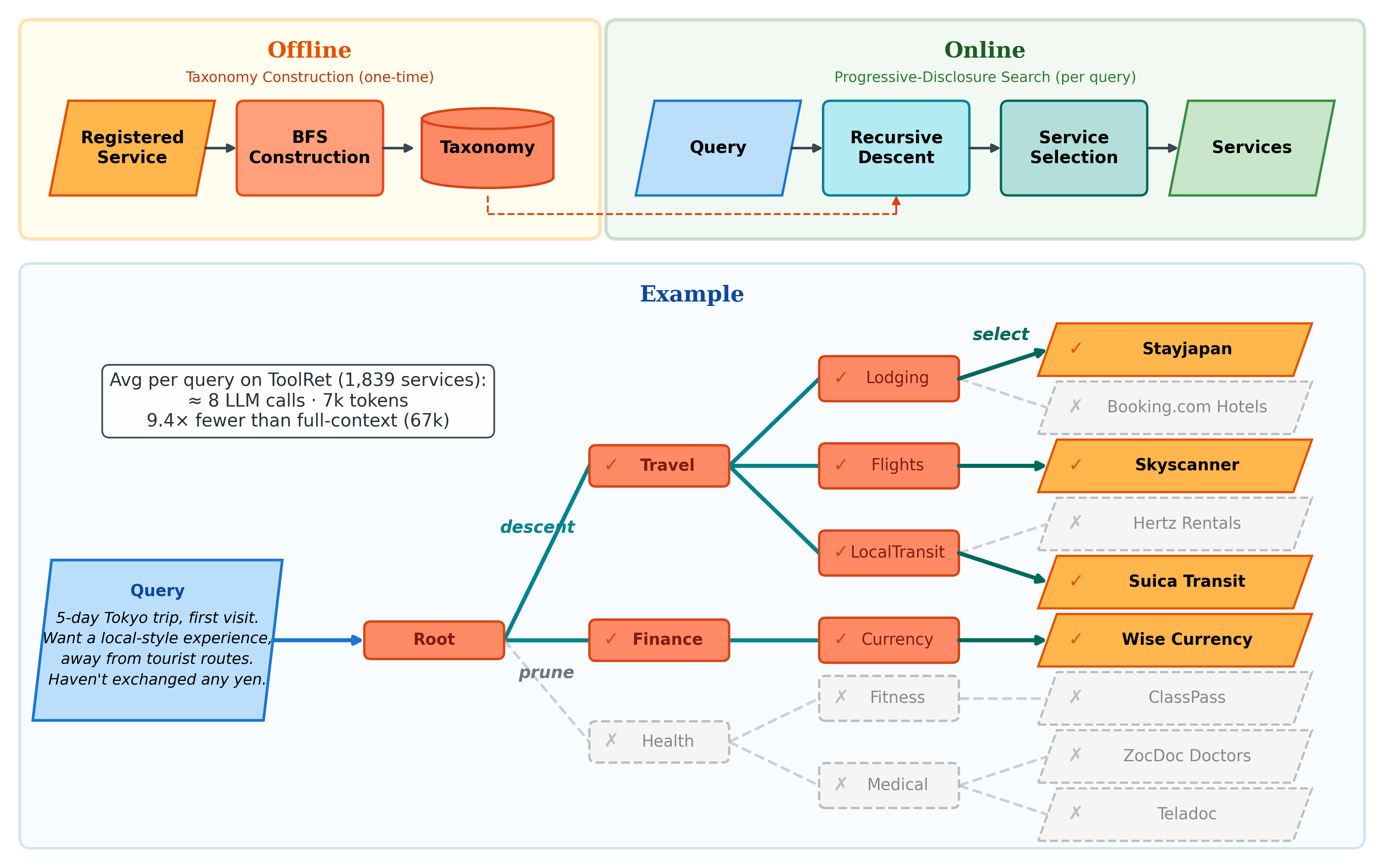}
\caption{A2X system overview. \textbf{Top-left}: an LLM automatically builds the hierarchical taxonomy from the registered services. \textbf{Top-right}: progressive disclosure recursively retrieves query-relevant categories and services. \textbf{Bottom}: LLM-native search over the taxonomy; solid colored arrows mark selected branches, gray dashed lines mark pruned ones.}
\label{fig:overview}
\end{figure*}

Realizing this approach comes with non-trivial implementation difficulty: manual taxonomy construction is costly, and one-shot LLM construction fails because the LLM cannot ingest thousands of services in a single context. We therefore deliver \textbf{A2X (Agent-to-Anything service discovery)} (Figure~\ref{fig:overview}): an LLM-native pipeline that constructs the taxonomy from raw service names and descriptions by recursive splitting, and at query time walks it via \emph{recursive descent} through the categories followed by \emph{service selection} at the reached leaves, producing a high-quality taxonomy with no human-curated ontology in the loop. By construction, A2X neutralizes both token bloat and Lost-in-the-Middle: every LLM call follows a short, focused recursive path and weighs only one node's children. Both index and retriever can be recomputed as the registry grows, and inherit multilingual capability from the underlying LLM. The full implementation is presented in \S\ref{sec:method}.

On the ToolRet benchmark (1{,}839 services, 1{,}714 queries), A2X reaches 92.6\,\% Hit Rate, beating full-context dumping by 6.2 absolute points at one-ninth the prompt-token cost and exceeding the state-of-the-art open-source embedding-based baseline by more than 20 Hit Rate points. The advantage carries to Chinese ToolRet, demonstrating cross-lingual robustness.

\paragraph{Contributions.} (i)~We reframe agent-side service discovery as a problem of context management and adopt LLM-native progressive disclosure as our approach to it (\S\ref{sec:problem}). (ii)~At the \emph{implementation} level, we deliver \textbf{A2X}, an LLM-native instantiation that automatically constructs the hierarchical taxonomy from raw service descriptions (\S\ref{sec:build}) and walks it layer by layer at query time (\S\ref{sec:search}), resolving the implementation difficulty that manual construction does not scale and one-shot LLM construction fails. (iii)~At the \emph{empirical} level (\S\ref{sec:experiments}), A2X strictly dominates standard baselines on Hit Rate and Recall on the ToolRet benchmark in both English and Chinese while consuming an order of magnitude fewer tokens than full-context dumping. (iv)~At the \emph{paradigm} level (\S\ref{sec:discussion}), we discuss LLM-native discovery as the natural endpoint of the trajectory along which inference cost continues to fall, and outline usage-aware refinement as a forward extension.

\section{Related Work}
\label{sec:related}

\paragraph{LLM context control.} Lost-in-the-Middle \citep{liu2024lostmiddle} and the RULER long-context benchmark \citep{hsieh2024ruler} establish that even when the input window is nominally large, models systematically under-attend to information placed in the middle of a long prompt: the binding constraint is \emph{effective} context rather than raw window size, and accuracy degrades smoothly as the prompt grows. A unifying response across LLM tasks is \emph{decomposition into local decisions}, which replaces a single global judgment over the full context with a sequence of focused local ones, so that each call inspects only a small portion of the candidate space. In multi-step reasoning, Chain-of-Thought \citep{wei2022cot} breaks a complex inference into intermediate steps, and Tree-of-Thoughts \citep{yao2023tot} organizes those steps into a search over partial-solution branches. Least-to-Most prompting \citep{zhou2023leasttomost} takes the idea further by first generating a list of ordered subproblems and then solving each in turn, so that the original question is never put to the LLM in one shot. In agent execution, ReAct \citep{yao2023react} interleaves reasoning traces with environment actions, so that each LLM call addresses only a small local subtask conditioned on accumulated observations rather than the entire trajectory. The shared pattern is to push attention away from a single long prompt and toward a sequence of short, focused ones.

\paragraph{Retrieval for service discovery.} Two paradigms dominate: \emph{LLM-only retrieval} and \emph{embedding-based retrieval}. The former, the standard MCP-client pattern, concatenates every service description into the prompt; token cost grows linearly with the registry, and both the context-window ceiling and Lost-in-the-Middle degradation make this route infeasible at scale. The latter, the current mainstream, replaces enumeration with dense vector similarity search backed by vector stores such as Chroma \citep{chroma2024}, and retrieval-augmented MCP frameworks such as RAG-MCP \citep{gan2025ragmcp} and MCP-Zero \citep{fei2025mcpzero}; however, fixed-dimensional embeddings struggle to capture the rich query-to-service semantic mapping an LLM can perform, and consequently miss relevant services: LiveMCPBench \citep{liu2025livemcpbench} attributes nearly half of MCP failures to the retrieval step. Beyond these two, a third line uses \emph{hand-curated hierarchical taxonomies} \citep{du2024anytool, agntcy2025dir}. These suffer from high maintenance cost and structural lag, making it difficult to keep pace with the rapidly growing and diversifying service population of the Internet-of-Agents era. In adjacent (non-service-discovery) domains, several LLM-assisted human-in-the-loop taxonomy-construction schemes have been proposed \citep{kargupta2025taxoadapt, golde2026hierarchical, shah2025intent}, but they rely on domain-specific prior knowledge and still require humans to provide or assist in constructing an initial taxonomy, which makes them hard to port directly to service discovery.

\section{Method}
\label{sec:method}

\subsection{Problem formulation}
\label{sec:problem}

A registry contains $N$ services $S = \{s_1, \dots, s_N\}$, each described by a name and a free-text natural-language description. Given a user query $q$, the goal is to return a small candidate set $\hat R \subseteq S$ that contains all services relevant to $q$.
The service-discovery problem is to realize a function $\hat R = \mathcal{F}(q, S)$, where $\mathcal{F}$ may invoke an LLM one or more times. Let $\tau(\mathcal{F}; q, S)$ denote the maximum prompt length any single LLM call inside $\mathcal{F}$ must process on input $(q, S)$. As established in \S\ref{sec:intro}, an LLM's effective context is bounded by some budget $B$ that is small relative to a full registry, even when the nominal window is large \citep{liu2024lostmiddle, hsieh2024ruler}. Reliable retrieval therefore requires $\tau(\mathcal{F}; q, S) \leq B$ for all $(q, S)$, with $B$ independent of $N$. The research question of this paper, \emph{context management for service discovery}, is how to design $\mathcal{F}$ satisfying the \textbf{context-management constraint}, while also maximizing retrieval accuracy and minimizing token consumption.

Our principal solution to this research question is \textbf{context isolation via progressive disclosure}: we realize the retrieval function $\mathcal{F}$ as a chain of local LLM calls $f_1, \dots, f_K$ acting on a shrinking sequence of candidate sets $V_1, \dots, V_K \subseteq S$, with $\hat R = (f_K \circ \dots \circ f_1)(q, S)$ and $|V_k| \leq B$ for every $k$, so that each call sees only the content under a single node of a hierarchical structure rather than any entire layer of $S$. The research question thereby decomposes into two concrete implementation problems. (i)~Manual taxonomy construction is high-cost and slow to update; how can an LLM construct the taxonomy automatically offline under the context-management constraint (\S\ref{sec:build})? (ii)~Given such a taxonomy, how do we use it efficiently at query time (\S\ref{sec:search})?

\subsection{Auto-constructed functional taxonomy}
\label{sec:build}

\begin{algorithm}[t]
\small
\caption{BFS Taxonomy Construction}
\label{alg:build}
\begin{algorithmic}[1]
\Require services $S$, $\theta_{kw}{=}500$, $\theta_{leaf}{=}40$, $D{=}3$
\State Create root node $r$ containing all of $S$; queue $Q \gets \{r\}$
\While{$Q$ non-empty}
  \State $v \gets Q.\text{pop}()$
  \If{$|v| > \theta_{kw}$}
    \State $K \gets$ \Call{BatchedKeywordExtract}{$v$} \Comment{$|v|/50$ LLM calls}
    \State $C \gets$ \Call{DesignFromKeywords}{$K$}
  \Else
    \State $C \gets$ \Call{DesignFromDescriptions}{$v$}
  \EndIf
  \If{$v = r$} $C \gets$ \Call{ValidateRoot}{$C$} \EndIf
  \Repeat \Comment{at most 3 iterations}
    \State assign each $s \in v$ to children $C$ in parallel
    \State collect $S_{gen}$ (matched ${>}|C|/3$ children), $S_{un}$ (matched no child)
    \If{$S_{gen} \cup S_{un} = \emptyset$} \textbf{break} \EndIf
    \State $C \gets$ \Call{Refine}{$C, S_{gen}, S_{un}$}
  \Until{converged or iteration cap}
  \State delete subcategories with $\le 2$ services
  \State enqueue children $c \in C$ if $|c| > \theta_{leaf}$ and $\text{depth}(c) < D$
\EndWhile
\State \Call{CrossDomainAssign}{$\text{leaves}$} \Comment{multi-parent phase}
\end{algorithmic}
\end{algorithm}

Algorithm~\ref{alg:build} summarizes the construction. The build is a Breadth-First Search (BFS) that starts from a single root containing every service. At each step we pop a node whose service count exceeds the leaf threshold $\theta_{leaf}$ and split it via a \emph{node splitter} that runs four substeps: (1)~category design, (2)~classification, (3)~refinement loop, (4)~edge handling. After BFS terminates, a separate cross-domain pass reassigns services that should live under multiple top-level domains. The paragraphs below describe each mechanism together with the build-time difficulty it addresses.

The three hyperparameters above admit short, principled defaults. $\theta_{kw} = 500$ marks the size above which packing service descriptions into one design prompt would risk context overflow and Lost-in-the-Middle degradation, so larger nodes route through the keyword-first compression described below. $\theta_{leaf} = 40$ matches the per-call service-selection budget; a node at or below this size is handled by a single leaf-level LLM call without further splitting. $D = 3$ reflects the empirical depth of real service catalogs (e.g., Travel $\rightarrow$ Flights $\rightarrow$ International); deeper trees increase build cost without recovering meaningful axes. All three are exposed as command-line flags and can be retuned for registries with different size and depth profiles.

\emph{Context overflow at large nodes} makes one-shot category design infeasible: concatenating $N{=}1{,}839$ descriptions at the root of ToolRet exceeds $200$k tokens and triggers Lost-in-the-Middle even when the window fits \citep{liu2024lostmiddle}. For nodes with more than $\theta_{kw}$ services we therefore use a \textbf{keyword-first} strategy: a \texttt{KeywordExtractor} processes services in batches of 50, extracting up to five functional keywords each, deduplicates across batches, and feeds a frequency-weighted keyword list to the designer. The resulting table is typically around 200 entries for $N{=}1{,}839$, compact enough to fit in a single design prompt; the designer never sees raw descriptions at scale, only an aggregated keyword distribution, keeping each prompt well under $B$.

\emph{Axis mixing} is the first design-time difficulty: when asked to organize a diverse service set, the LLM mixes orthogonal classification axes such as \emph{technology}, \emph{operation type}, or \emph{functional domain}, so that a flight-booking API could plausibly belong to ``Travel,'' ``API Services,'' and ``Transaction Processing'' simultaneously, destroying the tree's ability to prune. We address this with a \textbf{single-axis constraint}: the design prompt requires that all sibling categories within a node share one classification axis, and selects that axis along a preference chain. The chain prefers \emph{user-facing functional domains} such as ``Travel \& Tourism,'' ``Personal Finance,'' or ``Health \& Fitness''; if those fail to cover the node's services cleanly, it falls back to \emph{operation object}, then \emph{operation type}, and only as a last resort \emph{technical approach} such as ``AI Services'' or ``Cloud APIs.''

\emph{Sibling-boundary ambiguity} is the second difficulty: even within one axis, sibling categories can be semantically close, for instance ``Finance'' and ``Business,'' or ``Health'' and ``Fitness,'' leaving borderline services hard to disambiguate. We address this with an explicit \textbf{boundary clause} attached to every category in the design prompt: a ``NOT here'' statement that names what the category does \emph{not} cover, giving both the build-time classifier and the search-time navigator a clear disambiguation signal at every borderline.

Two run-time safety nets complement these design-time constraints. First, classifications may still match too many children or none at all; A2X therefore runs a \textbf{classification--refinement loop} in the spirit of \emph{harness engineering} \citep{hashimoto2026harness}, where services matching more than $1/3$ of the children flag unclear boundaries and services matching no child flag incomplete coverage, both fed back to the designer for a refined set of children. Second, a strict single-parent tree forces a binary choice for services that legitimately span multiple top-level domains and so harms recall; subcategories with two or fewer services are merged back into the parent, and a \texttt{CrossDomainAssigner} asks, for every leaf, ``which of these services should also appear under other top-level domains?'', placing accepted candidates in the best target leaf to give the taxonomy a \textbf{multi-parent structure}.

\subsection{Progressive-disclosure search}
\label{sec:search}

\begin{algorithm}[t]
\small
\caption{Progressive-Disclosure Retrieval}
\label{alg:search}
\begin{algorithmic}[1]
\Require query $q$, taxonomy $T$, mode $m$, $\theta_{merge}{=}30$
\Function{Navigate}{$v$, $q$}
  \State $C \gets \text{children}(v)$
  \If{$C = \emptyset$} \Return $\{(v, \text{services}(v))\}$ \EndIf
  \State $\hat C \gets$ LLM(\textsc{CategoryPrompt}$(m, q, C)$) \Comment{$\hat C \subseteq C$: selected children}
  \State \Return $\bigcup_{c \in \hat C}$ \Call{Navigate}{$c, q$} \Comment{parallel}
\EndFunction
\State $L \gets$ \Call{Navigate}{$\text{root}(T), q$}
\State $L \gets$ \Call{Deduplicate}{$L$} \Comment{first-come-first-served}
\State $G \gets$ \Call{MergeSmallGroups}{$L$, $\theta_{merge}$}
\State $R \gets \emptyset$
\ForAll{$g \in G$ \textbf{in parallel}}
  \State $R \gets R \cup$ LLM(\textsc{ServicePrompt}$(m, q, g)$)
\EndFor
\State \Return $R$
\end{algorithmic}
\end{algorithm}

Algorithm~\ref{alg:search} formalises the search procedure in two named steps. Parallel \textbf{recursive descent} (\texttt{CategoryNavigator}) walks the taxonomy top-down. At each node we present the children $C$ to the LLM as a numbered list of \texttt{name: description} pairs and ask for the indices of relevant children, denoted $\hat C \subseteq C$; the call is short, focused, and returns a comma-separated list. The selected children $\hat C$ are recursed into in parallel; the rest of the subtree is pruned and never inspected, and multiple selected nodes can be processed in parallel. \textbf{Service selection} (\texttt{ServiceSelector}) collects the leaf hits, deduplicates services across leaves (first-come), merges groups with fewer than $\theta_{merge}$ services using a least-common-ancestor distance heuristic to keep group sizes LLM-friendly, and finally issues one parallel LLM call per merged group to pick the relevant services.

The same algorithm supports three modes that differ only in the system prompts. Quoting from our implementation: \texttt{get\_all} asks the LLM to ``select all categories that could contain query-relevant services,'' tuned for high recall; \texttt{get\_important} asks the LLM to ``select all categories that could contain query-relevant services, but deduplicate services with the same function''; \texttt{get\_one} asks the LLM to ``always select the single most relevant branch.''

The number of LLM calls per query is approximately $d \cdot b + g$, where $d$ is the average navigation depth, $b$ is the average branching factor \emph{selected} per level (typically much smaller than the taxonomy's branching factor $B$), and $g$ is the number of merged leaf groups visited during service selection. Under a balanced taxonomy with branching factor $B$, $d = \Theta(\log_B N)$; both $b$ and $g$ depend on query specificity rather than on $N$ -- empirically (\S\ref{sec:study1}) they are bounded by small constants on ToolRet. The total prompt-token cost per query therefore \textbf{grows sub-linearly in $N$} under registry growth that preserves the branching structure; we do not claim an asymptotic $O(\log N)$ bound, since the constants in $b$ and $g$ are data-dependent. A per-$N$ scaling characterization across $10^2$--$10^5$ is left to future work (\S Limitations).

\begin{figure*}[t]
\centering
\includegraphics[width=0.736\textwidth]{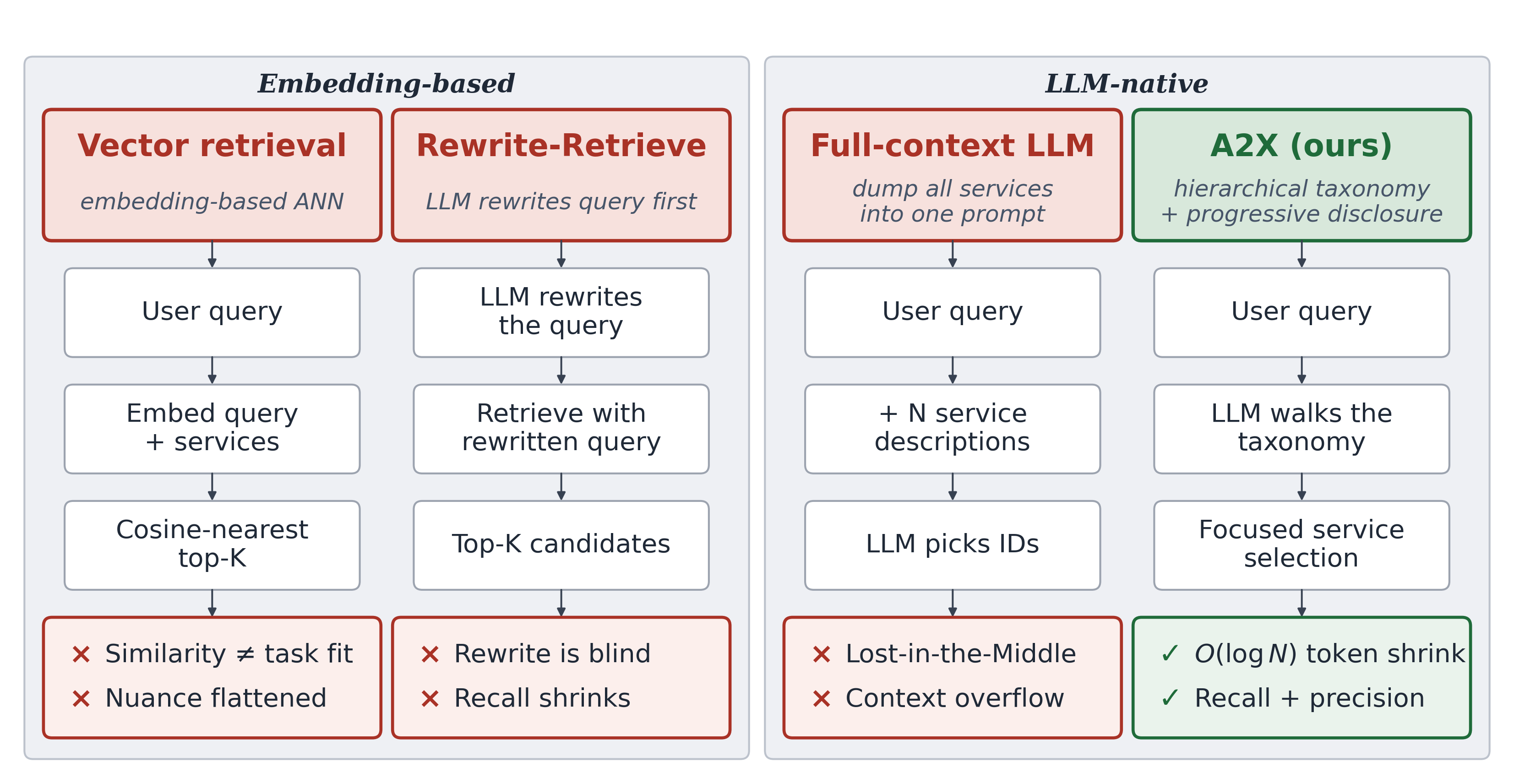}
\caption{\textbf{Service-discovery method comparison.} \emph{Embedding-based} methods (columns 1--2: vector approximate-nearest-neighbor (ANN) retrieval and rewrite-retrieve) and the naive \emph{LLM-native} baseline (column 3: full-context LLM) each suffer a distinct failure mode that surfaces as insufficient recall or context overflow. A2X (column 4) is also LLM-native, but pairs a hierarchical taxonomy with progressive disclosure, turning service discovery into a sequence of short, focused LLM decisions that achieve sub-linear token cost in $N$ (full cost model in \S\ref{sec:study1}) while preserving recall and precision.}
\label{fig:methods}
\end{figure*}

\section{Experiments}
\label{sec:experiments}

\subsection{Datasets and metrics}
\label{sec:datasets}

We use a cleaned variant of ToolRet \citep{shi2025toolret} as our tool-retrieval benchmark, filtering out low-quality entries (e.g., ``the tool that opens the navigation bar,'' which even a human reader cannot interpret without context). The cleaned variant contains 1{,}839 services and 1{,}714 queries. We additionally use ToolRet\_CN, a Chinese version of the same benchmark with the same service inventory and query set, to test cross-lingual robustness. For the cross-dataset comparison in \S\ref{sec:embedding}, we additionally evaluate on \emph{publicMCP} (1{,}387 publicly available MCP servers from \url{https://github.com/modelcontextprotocol/servers} paired with 50 queries) and its Chinese counterpart publicMCP\_CN. Datasets are released alongside this paper to support future work on tool retrieval.

We report Hit Rate and Recall. Hit Rate measures whether the returned set contains at least one ground-truth relevant service; Recall is the fraction of relevant services that are returned; Precision (reported in the appendix only) is the fraction of returned services that are relevant. We deliberately do not report Precision as a primary metric: large-scale tool-retrieval benchmarks face an inherent annotation challenge where functionally equivalent services are hard to label exhaustively, so the measured Precision systematically underestimates the true relevance ratio; the consequences are discussed further in \S Limitations. The remainder of \S\ref{sec:experiments} reports three comparison studies that isolate the search-side gain (\S\ref{sec:study1}), the build-side gain (\S\ref{sec:study2}), and the gain over embedding-based retrieval together with a cross-lingual evaluation (\S\ref{sec:embedding}). All baselines use the same underlying LLM (DeepSeek-V3.2) where applicable.

\subsection{Study 1: A2X search vs pure-LLM full-context retrieval}
\label{sec:study1}

This study isolates A2X's search-side contribution. With the taxonomy already built, we compare A2X's progressive-disclosure search against a \emph{pure-LLM} baseline that concatenates every service description into a single prompt and asks the LLM to return relevant service IDs. Tool input schemas are excluded from this baseline's catalog payload, matching A2X's setting; including them would raise the per-query input to approximately 204k tokens.

\begin{table}[t]
\centering
\small
\setlength{\tabcolsep}{5pt}
\begin{tabular}{lrrr}
\toprule
\textbf{Method} & \textbf{Tok/q} & \textbf{HR $\uparrow$ (\%)} & \textbf{Recall $\uparrow$ (\%)} \\
\midrule
Pure-LLM & 67{,}619 & 86.4 & 84.7 \\
\textbf{A2X (\texttt{get\_all})} & \textbf{7{,}069} & \textbf{92.6} & \textbf{89.2} \\
\bottomrule
\end{tabular}
\caption{\textbf{Search-side comparison on ToolRet (1{,}839 services, 1{,}714 queries).} Both methods are evaluated on the full 1{,}714-query test set. A2X improves Hit Rate by 6.2 absolute points and Recall by 4.5 points over pure-LLM full-context retrieval, while consuming roughly $1/9$ the prompt tokens.}
\label{tab:study1}
\end{table}

The pure-LLM baseline scans all 1{,}839 services in a single LLM call, exposing it to Lost-in-the-Middle effects \citep{liu2024lostmiddle}. A2X instead makes a sequence of focused decisions over short candidate lists: each navigation step presents only about 8 category options to the LLM (vs.\ 1{,}839 services for the baseline), and each leaf-group call presents about 12 services after recursive-descent pruning. Following the cost model in \S\ref{sec:search}, the per-query LLM-call count is $d \cdot b + g$, where $d = \Theta(\log_B N)$ under a balanced taxonomy of branching $B$, while $b$ (selected branches per level) and $g$ (visited leaf groups) depend on query specificity rather than $N$. The per-query token cost is therefore $(d \cdot b + g) \cdot c$ where $c$ is the per-call prompt budget bounded by the LLM's effective context window. This grows sub-linearly in $N$: empirically on ToolRet ($N = 1{,}839$) A2X averages 8.0 LLM calls and 7{,}069 tokens per query, a $9\times$ reduction over full-context dumping's 67{,}619 tokens, alongside the 6.2-point hit-rate gain.

\subsection{Study 2: Contribution of each A2X build module}
\label{sec:study2}

This study isolates A2X's build-side contribution. A2X's BFS recursive construction issues many short, focused LLM calls. We compare it against a pure-LLM baseline, \emph{one-shot build}: the LLM reads all service descriptions in a single call and emits the taxonomy skeleton, after which parallel LLM calls assign each service to one of its leaves. We then evaluate downstream search on the resulting taxonomy using the same A2X search procedure as Study~1, so any quality difference traces back to the taxonomy itself. We additionally test the effect of each build module individually. Adding any one module to the one-shot baseline lifts Recall by 7.8 to 12.2 absolute points, with the refinement loop the strongest single contributor (+12.2 R) and the keyword frequency table a close second (+11.7 R) at less than half its token cost. Enabling all three modules and combining them with recursive BFS subdivision closes the remaining gap to 89.2 Recall.

\begin{table}[t]
\centering
\small
\setlength{\tabcolsep}{4pt}
\begin{tabular}{lcc}
\toprule
\textbf{Build method} & \textbf{HR $\uparrow$ (\%)} & \textbf{Recall $\uparrow$ (\%)} \\
\midrule
One-shot (baseline)              & 59.7 & 54.6 \\
\;+ Freq table                    & 70.8 & 66.3 \\
\;+ Refine loop                   & 71.5 & 66.8 \\
\;+ Single-axis                   & 66.6 & 62.4 \\
\textbf{A2X recursive BFS (all)} & \textbf{92.6} & \textbf{89.2} \\
\bottomrule
\end{tabular}
\caption{\textbf{Build-side comparison and additive module ablation on ToolRet (1{,}839 services, 1{,}714 queries).} Search HR and Recall reported after running A2X search (\texttt{get\_all} mode) over each taxonomy.}
\label{tab:study2}
\end{table}

\subsection{Study 3: A2X vs embedding-based retrieval}
\label{sec:embedding}

This study compares A2X against the dominant retrieval paradigm of dense vector similarity search. We test four open-source embedding baselines (MiniLM, BGE-large-en, text2vec-base-chinese for the Chinese setting, and the multilingual BGE-M3), plus the LLM-query-rewriting baseline MCP-Zero \citep{fei2025mcpzero}. All embedding baselines rank candidates by cosine similarity to the query; MCP-Zero uses BGE-M3 as its retrieval backbone. We evaluate on both datasets (introduced in \S\ref{sec:datasets}) and their Chinese counterparts (ToolRet\_CN / publicMCP\_CN) for cross-lingual robustness.

\begin{table*}[t]
\centering
\small
\setlength{\tabcolsep}{6pt}
\begin{tabular}{llrrrr}
\toprule
\textbf{Dataset} & \textbf{Method} & \multicolumn{2}{c}{\textbf{EN}} & \multicolumn{2}{c}{\textbf{CN}} \\
\cmidrule(lr){3-4} \cmidrule(lr){5-6}
                & & HR $\uparrow$ (\%) & Recall $\uparrow$ (\%) & HR $\uparrow$ (\%) & Recall $\uparrow$ (\%) \\
\midrule
\multirow{6}{*}{ToolRet}   & MiniLM                & 69.1 & 61.8 & --   & --   \\
                           & BGE-large-en          & 68.0 & 61.7 & --   & --   \\
                           & text2vec-base-chinese & --   & --   & 41.4 & 36.9 \\
                           & BGE-M3                & 64.8 & 58.0 & 60.9 & 54.4 \\
                           & MCP-Zero              & 62.1 & 55.9 & 58.6 & 52.4 \\
                           & \textbf{A2X}          & \textbf{92.6} & \textbf{89.2} & \textbf{90.3} & \textbf{86.4} \\
\midrule
\multirow{6}{*}{publicMCP} & MiniLM                & 80.0 & 49.8 & --   & --   \\
                           & BGE-large-en          & 70.0 & 40.5 & --   & --   \\
                           & text2vec-base-chinese & --   & --   & 28.0 & 11.4 \\
                           & BGE-M3                & 70.0 & 41.8 & 74.0 & 43.3 \\
                           & MCP-Zero              & 52.0 & 27.5 & 48.0 & 23.3 \\
                           & \textbf{A2X}          & \textbf{100.0} & \textbf{94.9} & \textbf{98.0} & \textbf{88.9} \\
\bottomrule
\end{tabular}
\caption{\textbf{Comparison against embedding-based retrieval on ToolRet and publicMCP.} Embedding baselines return the top-$K$ cosine-nearest candidates; here we show $K{=}5$ for ToolRet (mean 1.34 ground-truth services per query) and $K{=}10$ for publicMCP (mean 2.78). Full $K$ sweep in Appendix~\ref{sec:appendix_embed}.}
\label{tab:study3}
\end{table*}

Compared to LLM-native approaches, embedding-based methods struggle to capture the complex query-service semantic mapping: even the strongest open-source embedding in our lineup still trails A2X by 23.5 HR and 27.4 Recall points on EN (MiniLM), and by 29.4 HR / 32.0 Recall on CN (BGE-M3). We additionally test MCP-Zero with the prompt template released by \citet{fei2025mcpzero}, which underperforms its retrieval backbone (BGE-M3) by 2.7 HR on our benchmark. The embedding-based family thus shares one bottleneck, namely insufficient recall. The LLM-native A2X sidesteps this paradigm by recasting service discovery as a sequence of discrete LLM decisions over short, focused candidate lists drawn from a taxonomy of functional categories, while also interpreting the various positive and negative preferences a user may attach to a query. A2X also achieves accuracy on the Chinese dataset comparable to English, demonstrating its cross-lingual strength.

\section{Discussion}
\label{sec:discussion}

The core insight behind A2X is a division of labor that plays to the LLM's strengths while avoiding its weaknesses. LLMs excel at the kind of semantic reasoning service discovery demands, mapping a vague goal such as ``plan a trip to Tokyo'' to functional categories like \emph{Flights}, \emph{Hotels}, and \emph{Currency}, and disambiguating positive and negative preferences in the query. They degrade, however, when forced to scan thousands of candidates in a single pass, the Lost-in-the-Middle effect that gives rise to the context-management constraint of \S\ref{sec:problem}. The taxonomy resolves this tension by interposing between query and registry: every LLM call sees only one node's children, never more than a dozen options, and irrelevant branches are pruned before they reach the model. This same design choice, \emph{context isolation via progressive disclosure}, underlies all three results in \S\ref{sec:experiments}: the search-side gain, the build-side gain, and the gap against the strongest dense embeddings. A single mechanism thus addresses both the LLM-context bottleneck and the recall ceiling that limits embedding-based discovery.

The contrast with embedding-based retrieval is paradigmatic rather than incremental. Dense retrievers compress both query and service into fixed-dimensional vectors and rank by cosine similarity, leaving little room for fine-grained intent or domain-specific tool vocabulary that an LLM can still parse from the raw text. Hierarchical alternatives such as AnyTool \citep{du2024anytool} depend on a human-curated category structure that ages out as the registry grows. A2X discards both assumptions: each navigation step is a discrete LLM decision rather than a vector-similarity score, and it operates on candidate lists drawn from a taxonomy that the LLM builds autonomously. The cross-lingual robustness in \S\ref{sec:embedding} is a concrete consequence of the same design: language coverage is inherited from the underlying LLM rather than from per-language embedding engineering, so the discovery layer evolves with the LLM itself rather than with a separate retrieval stack.

\section{Conclusion}
\label{sec:conclusion}

We presented \textbf{A2X (Agent-to-Anything service discovery)}, an LLM-native service-discovery system that reframes the registry as a problem of \emph{context management for service discovery} and addresses it through \emph{context isolation via progressive disclosure}: an LLM-built hierarchical taxonomy is walked layer by layer at query time, so that every LLM call sees only one node's children rather than the whole registry. The pipeline is fully autonomous: an LLM builds the taxonomy directly from raw service descriptions, and an LLM navigates it at query time via \emph{recursive descent} through the categories followed by \emph{service selection} at the reached leaves. On ToolRet, A2X reaches 92.6\,\% Hit Rate in English and 90.3\,\% in Chinese at roughly one-ninth the prompt-token cost of full-context dumping, strictly dominating the full-context LLM baseline by 6.2 absolute HR points and exceeding the strongest open-source embedding-based baseline by more than 20 HR points.

Looking forward, the LLM-native discovery layer scales naturally in two directions. At Internet scale it can act as a unified discovery layer across heterogeneous services, agents, and protocol entry points, providing what amounts to an \emph{agent DNS} across organizational and platform boundaries; inside an organization, the same mechanism becomes a governable service catalog over internal tools, endpoints, and workflow nodes. Because every navigation step is an LLM decision rather than a vector lookup, users can encode richer constraints in the query, namely budgets, exclusions, and negative preferences, while long-tail and specialized service providers gain reliable discoverability without depending on a small number of traffic platforms. As the agent ecosystem grows toward registries an order of magnitude larger than today's benchmarks, we therefore view LLM-native discovery not as a niche alternative to embedding-based retrieval but as its natural successor.

\section*{Limitations}

\paragraph{Static taxonomy.} Our build pipeline treats services as static and is invariant to query frequency. A natural next step is \emph{usage-aware refinement}: once a deployment collects service-usage logs, the taxonomy can be analyzed under a minimum-cost objective (expected query cost $= \sum_s \text{frequency}(s) \times \text{depth}(s)$) to identify high-frequency services placed at deep nodes and long-tail services placed at shallow nodes; the resulting signal can be fed back to the LLM as a soft hint to lift frequently queried services to shallower positions and push long-tail services deeper, shortening the expected navigation path.

\paragraph{Language coverage.} Our cross-lingual evaluation (\S\ref{sec:embedding}) covers machine-translated Chinese versions of both ToolRet and publicMCP and shows strong results. However, the Chinese data is machine-translated, not natively authored. Performance on natively non-English registries, particularly languages where service descriptions and queries draw on incompatible vocabularies, may require taxonomy rebuilds rather than translation, and remains to be studied.

\paragraph{LLM dependency.} Both the build pipeline and the retrieval algorithm rely on a capable underlying LLM. Our experiments use a cost-efficient commercial endpoint (DeepSeek-V3.2 via the official API); a substantially weaker model would likely degrade taxonomy quality and, in turn, retrieval accuracy. We have not characterised this degradation curve.

\paragraph{Precision under-reporting.} As discussed in \S\ref{sec:datasets}, ground-truth annotations in tool-retrieval benchmarks are systematically incomplete, so traditional Precision metrics under-report the quality of high-recall methods. The pattern is visible in Table~\ref{tab:appendix_prec_e3} (Appendix~\ref{sec:appendix_embed}): A2X's Precision sits within $\pm$3 points of the strongest embedding baseline on each dataset (ToolRet 16.2 / 12.0 vs.\ 15.3 / 13.5; publicMCP 10.5 / 9.9 vs.\ 13.0 / 11.8), while A2X's Recall leads by 27--32 points on ToolRet and 45+ points on publicMCP. Under incomplete labels, the high-Recall method correctly retrieves more services that are functionally relevant but unlabeled, and each such item is counted as a false positive against it. We therefore read the Precision gap as label-coverage noise rather than evidence of weaker selection, but it does limit our ability to make strong precision claims against prior work that reports Precision-based scores.

\paragraph{Build cost at registry scale.} Our build experiments target registries of $\sim$2k services. Larger registries (tens of thousands of services, multiple updates per day) will exercise the incremental-rebuild path of the system, which is implemented in prototype form but not yet benchmarked at scale.

\section*{Ethics Statement}

A2X is a retrieval algorithm and does not generate new content or take autonomous actions. The datasets used in this work (ToolRet and publicMCP, each in English and Chinese) are public collections of API and tool descriptions; they do not contain personal data. Our LLM API usage went through standard commercial endpoints under their published terms of service; we logged no personally identifiable information. Service registries can amplify whichever services are indexed, and a deployed A2X instance inherits the curation choices of its maintainers; we encourage operators of agent registries to publish their indexing policies.

\bibliography{custom}
\bibliographystyle{acl_natbib}

\input{appendix}

%% file: appendix.tex
\appendix

\section*{Appendix Contents}
\begin{itemize}
\setlength{\itemsep}{0pt}
\item Appendix~\ref{sec:appendix_impl}: Implementation Details (hyperparameters, models, reproducibility).
\item Appendix~\ref{sec:appendix_cleaning}: Datasets (ToolRet cleaning protocol; publicMCP construction).
\item Appendix~\ref{sec:appendix_taxonomy}: Taxonomy Structure (taxonomy size statistics; branching factor discussion).
\item Appendix~\ref{sec:appendix_results}: Additional Experimental Results (search modes, build cost, DeepSeek V4 robustness check).
\item Appendix~\ref{sec:appendix_baselines}: Baseline Implementation Details (pure-LLM, one-shot build, embedding baselines, MCP-Zero).
\end{itemize}

\section{Implementation Details}
\label{sec:appendix_impl}

\paragraph{Build hyperparameters.} We use \texttt{keyword\_threshold}$=$500, \texttt{max\_service\_size}$=$40, \texttt{max\_depth}$=$3, \texttt{generic\_ratio}$=$1/3, \texttt{max\_categories\_size}$=$20, \texttt{max\_refine\_iterations}$=$3, \texttt{keyword\_batch\_size}$=$50. Parallelism is set to 20 worker threads.

\paragraph{Search hyperparameters.} Search uses the leaf-group merge threshold $\theta_{merge}$. Modes \texttt{get\_all}, \texttt{get\_important}, \texttt{get\_one} share the same machinery and differ only in their prompt instructions (quoted in \S\ref{sec:search}).

\paragraph{Models.} All LLM calls in build, search, and the one-shot baseline use a chat-completion endpoint at temperature $0$. We use a cost-efficient commercial endpoint (DeepSeek-V3.2); the algorithm is model-agnostic. Embedding baselines use, in increasing capacity: \texttt{sentence-transformers/all-MiniLM-L6-v2} (384d), \texttt{BAAI/bge-large-en-v1.5} (1{,}024d), and the multilingual \texttt{BAAI/bge-m3} (1{,}024d). For the Chinese-only open-source baseline we use \texttt{shibing624/text2vec-base-chinese} (768d).

\paragraph{Reproducibility.} The full build and search code, the prompt templates, the dataset processing scripts, the one-shot baseline, the MCP-Zero baseline, and the evaluation harness are released with the paper. Each experiment in \S\ref{sec:experiments} has its own output directory under \texttt{results/}, containing a \texttt{summary.json} and a \texttt{per\_query.jsonl} for full audit.

\section{Datasets}
\label{sec:appendix_cleaning}

\subsection{ToolRet (cleaned)}

The original ToolRet collection \citep{shi2025toolret} aggregates 44{,}453 tools from a dozen upstream sources (\texttt{toolACE}, \texttt{toolbench}, \texttt{apigen}, \texttt{toolalpaca}, \texttt{ultraTool}, \texttt{gorilla}, \texttt{toolink}, \texttt{toolLens}, \texttt{reversechain}, \texttt{webtools}, \texttt{metatool}, and others). Across these sources the per-tool documentation has very uneven quality: some entries are well-described public APIs, others are unlabeled lambda functions or near-duplicate auto-generated stubs. Because A2X organizes services by their \emph{user-facing functional description}, the build pipeline is sensitive to description quality: a service whose description is too short, too generic, or too platform-specific to be self-contained will degrade both the build (the LLM cannot reliably categorize it) and the search (the LLM cannot reliably distinguish it from its neighbors). To obtain a reliable benchmark we therefore curated the cleaned variant used throughout the paper.

\paragraph{Process.} The reviewer was shown the candidate tools alongside the query set, and was asked to flag any tool that triggered one of the four exclusion criteria below. The reviewer's flags were then reconciled with the query coverage and merged into the kept set. The cleaned dataset is released, the rejected entries can be diffed against the original by \texttt{id}, and the criteria are stated below in full.

\paragraph{Exclusion criteria.}
\begin{enumerate}
\itemsep0pt
\item \textbf{No corresponding query.} The original ToolRet contains many tools that no query in the benchmark ever points at. Retaining them inflates the registry without affecting any HR / Recall measurement; we drop tools with zero ground-truth coverage.
\item \textbf{Near-duplicate of another retained tool.} Several upstream sources are dense with auto-generated variants of the same underlying API (e.g., paginated read endpoints, locale-shifted clones). When two tools have the same effective function and roughly interchangeable descriptions, we keep one representative.
\item \textbf{Description too short.} We drop tools whose description is a single fragment of fewer than roughly fifteen tokens, which is below the floor at which any classifier (human or LLM) can place the tool with confidence. Example removed: \texttt{toolbench\_tool\_4327}, ``Retrieves an order''.
\item \textbf{Description not self-contained or not human-interpretable.} We drop tools whose description requires substantial out-of-band context to understand. Two recurring patterns: (a) platform-specific action stubs (e.g., ``the tool that opens the navigation bar'', a description that is meaningful only inside one fixed UI and to a reader familiar with it); (b) lambdas whose name and one-line docstring are opaque without seeing the body (e.g., \texttt{toolink\_tool\_1123}, ``\texttt{goats\_per\_cow}: To calculate the number of goats equivalent to one cow'', which the original code reveals to mean a unit-conversion expression that returns the constant 21/15).
\end{enumerate}

\paragraph{Outcome.} The cleaned variant retains 1{,}839 tools, a 4.1\% subset of the original ToolRet, paired with 1{,}714 English queries whose ground-truth sets are entirely contained in the retained tools (mean ground-truth set size 1.34 services per query). Coverage is preserved across the major upstream sources (\texttt{toolACE}, \texttt{apigen}, \texttt{toolbench}, \texttt{reversechain}, \texttt{toolLens}, \texttt{ultraTool}, \texttt{webtools}, \texttt{metatool}, \texttt{toolalpaca}, with the original mix's long tail still represented). We additionally publish a Chinese variant, ToolRet\_CN, produced by translating each retained tool's name and description and the matching queries while preserving the ground-truth mapping. Both variants are released alongside this paper.

\subsection{publicMCP}
\label{sec:appendix_publicmcp}

\emph{publicMCP} is a 1{,}387-entry catalog of publicly available MCP server implementations that we assembled to test cross-dataset robustness on a registry whose composition differs from ToolRet's. The catalog draws from the official MCP servers repository,\footnote{\url{https://github.com/modelcontextprotocol/servers}} which its own README characterizes as \emph{``a collection of reference implementations for the Model Context Protocol (MCP), as well as references to community built servers and additional resources.''} Each entry consists of a name, a short functional description (median 115 characters, mean 129), and a URL to the source repository (91.6\% on GitHub, with the remainder split across npm, GitLab, and project documentation pages).

\paragraph{Queries and ground truth.} The catalog is paired with 50 English natural-language queries describing concrete agent tasks (e.g., \textit{``open a pull request and notify the engineering channel''}). A Chinese counterpart \emph{publicMCP\_CN} is produced by translating both the service descriptions and the queries while preserving the ground-truth mapping.

\section{Taxonomy Structure}
\label{sec:appendix_taxonomy}

Table~\ref{tab:taxonomy_stats} reports the structural properties of the automatically constructed taxonomies on the English and Chinese versions of ToolRet. A2X rebuilds the taxonomy independently for each language directly from the service descriptions; no cross-lingual alignment is performed.

\begin{table}[t]
\centering
\small
\begin{tabular}{lrr}
\toprule
\textbf{Statistic} & \textbf{ToolRet} & \textbf{ToolRet\_CN} \\
\midrule
Total categories     & 352  & 358 \\
Leaf categories      & 324  & 328 \\
Max depth            & 3    & 3 \\
Avg services / leaf  & 12.7 & 12.5 \\
\bottomrule
\end{tabular}
\caption{Taxonomy structure statistics on ToolRet (English) and ToolRet\_CN (Chinese). Sizes are nearly identical, reflecting that A2X inherits its multilingual reasoning from the underlying LLM rather than from any language-specific component.}
\label{tab:taxonomy_stats}
\end{table}

\paragraph{On branching factor.}
Information-theoretically, a binary tree ($b{=}2$) minimizes the maximum number of decisions to reach any leaf: $\lceil\log_2 N\rceil$ levels for $N$ services. However, three practical considerations lead us to prefer a moderate branching factor ($b \in [5, 20]$). First, every LLM call carries a fixed-cost overhead (system prompt, query, instruction template) regardless of how many candidates are presented; a node with 10 to 15 children amortizes this overhead across more useful decisions per call. Second, forced binary splits often produce unnatural partitions (e.g., ``Domestic Travel'' vs.\ ``Everything Else''), degrading classification accuracy. Third, a shallower tree with moderate branching reduces the total number of sequential LLM calls: a depth-3 tree with branching factor 15 covers $15^3 = 3{,}375$ leaf slots in 3 calls, whereas a binary tree needs $\lceil\log_2 3375\rceil = 12$ calls. Our empirical configuration ($b \in [5, 20]$, $D{=}3$) achieves an average of 7.96 calls per query on ToolRet.

\section{Additional Experimental Results}
\label{sec:appendix_results}

\subsection{Search modes}
\label{sec:appendix_search_modes}

The same A2X search algorithm supports three modes that differ only in the navigation system prompts and that sweep a smooth precision-recall trade-off. Table~\ref{tab:modes} compares them on ToolRet.

\begin{table}[t]
\centering
\small
\begin{tabular}{lrrr}
\toprule
\textbf{Mode} & \textbf{Recall $\uparrow$ (\%)} & \textbf{Tokens/q} & \textbf{Calls/q} \\
\midrule
\texttt{get\_one}      & 42.99 & 3{,}648 & 3.42 \\
\texttt{get\_important}& 74.51 & 5{,}210 & 5.71 \\
\texttt{get\_all}      & 89.19 & 7{,}069 & 7.96 \\
\bottomrule
\end{tabular}
\caption{Search modes on ToolRet. The same taxonomy and the same algorithm are reused; only the per-call prompt instructions differ.}
\label{tab:modes}
\end{table}

\texttt{get\_one} runs at less than half the cost of \texttt{get\_all} and is the natural choice when the agent needs the single best tool; \texttt{get\_all} maximizes recall when the agent will downstream-rerank. Varying only the navigation prompts (while keeping the same taxonomy fixed) sweeps a smooth precision-recall curve, evidence that the taxonomy is genuinely the structural object the LLM is reasoning over.

\subsection{Build cost analysis}
\label{sec:appendix_build_cost}

The taxonomy is built \emph{once}, fully automatically and without human intervention, and reused for all subsequent queries. For ToolRet ($N{=}1{,}839$), the build issues 18{,}206 LLM calls consuming $\sim$29M tokens and completes in approximately 3 hours with 20 parallel workers on a commodity endpoint (DeepSeek-V3.2), costing roughly \$4 to \$8 at current API rates. The BFS splitting phase accounts for 92\% of the cost (16{,}797 calls); the cross-domain phase adds only 1{,}409 calls.

To put this in perspective: serving 4{,}100 queries under the pure-LLM full-context baseline would consume $4{,}100 \times 67{,}619 \approx 277$\,M tokens, roughly \textbf{9.6$\times$ the entire build budget}, while achieving lower recall. A2X's one-time build cost is therefore recouped in fewer than 4{,}100 queries, after which every additional query is $9\times$ cheaper than the full-context alternative.

\subsection{DeepSeek V4 robustness check}
\label{sec:appendix_v4}

DeepSeek V4 was released in April 2026 as the next-generation successor to V3.2, in two tiers: \texttt{deepseek-v4-flash} (284B parameters, 13B active) and \texttt{deepseek-v4-pro} (1.6T parameters, 49B active). Both support the same OpenAI-compatible endpoint as V3.2 and default to thinking mode on; we disable thinking via the request-body flag \texttt{"thinking": \{"type": "disabled"\}} for every LLM call, since our classification and selection prompts do not need a reasoning trace and the trace would inflate output tokens. To verify A2X is not tied to one specific model, we fully rebuild the taxonomy from raw service descriptions under each V4 generation (using both the recursive BFS build and the one-shot baseline) and replay both the search-side (Study~1) and build-side (Study~2) comparisons.

\begin{table*}[h]
\centering
\small
\setlength{\tabcolsep}{6pt}
\begin{tabular}{llrrrr}
\toprule
\textbf{Setup} & \textbf{Model} & \textbf{Tok/q} & \textbf{HR $\uparrow$ (\%)} & \textbf{Recall $\uparrow$ (\%)} & \textbf{Prec. (\%)} \\
\midrule
Pure-LLM             & V3.2     & 67{,}619        & 86.4          & 84.7          & 16.8 \\
                     & V4-flash & 68{,}317        & 94.0          & 91.3          & 9.8 \\
                     & V4-pro   & 67{,}688        & 86.0          & 82.3          & 14.9 \\
\midrule
A2X (recursive BFS)  & V3.2     & 7{,}069         & 92.6          & 89.2          & 16.2 \\
                     & V4-flash & 8{,}088         & \textbf{94.3} & \textbf{91.9} & 10.6 \\
                     & V4-pro   & \textbf{5{,}405}& 89.7          & 85.8          & \textbf{23.8} \\
\midrule
A2X (one-shot)       & V3.2     & 3{,}456         & 59.7          & 54.6          & 12.0 \\
                     & V4-flash & 3{,}851         & 61.6          & 56.8          & 11.7 \\
                     & V4-pro   & \textbf{3{,}096}& \textbf{70.5} & \textbf{65.7} & \textbf{22.4} \\
\bottomrule
\end{tabular}
\caption{Search-side and build-side comparisons replayed under three DeepSeek model generations. All A2X rows are on the full 1{,}714-query ToolRet test set. The V3.2 pure-LLM row is also on the full 1{,}714 queries (matching Table~\ref{tab:study1}); the V4-flash and V4-pro pure-LLM rows are still on the 50-query subsample and will be re-run on the full set in a future revision. Bold values are the best within each block.}
\label{tab:v4-robustness}
\end{table*}

\paragraph{Search-side observations.} \textbf{(i) A2X is model-stable on token cost.} Across V3.2 and both V4 tiers, A2X (recursive build) consumes 5{,}405--8{,}088 tokens per query, never more than 12\% of the corresponding full-context cost on the same model. The roughly $9\times$ token-efficiency lead is invariant to model generation. \textbf{(ii) Long-context skill compresses A2X's accuracy lead.} V4-flash is unusually strong on long-context: its full-context HR jumps from 86.4 (V3.2) to 94.0, partially neutralizing the Lost-in-the-Middle effect at the 68k-token registry scale. A2X is still slightly ahead on accuracy here (94.3 vs.\ 94.0), but the bulk of A2X's advantage on V4-flash is now cost rather than accuracy. The accuracy gap reopens on V4-pro (89.7 vs.\ 86.0, +3.7 HR points) and was largest on V3.2 (92.6 vs.\ 86.4, +6.2). \textbf{(iii) V4-pro trades recall for precision; A2X passes through the model's selection profile.} V4-pro builds a tighter taxonomy than V4-flash (212 leaves vs.\ 344) and returns fewer but more precise services per query (A2X precision 23.8 vs.\ flash's 10.6, at the cost of 6 recall points). This is a model-level preference, not a method-level limitation; A2X's algorithmic structure is invariant and downstream Hit Rate remains in the 90\% range across all three generations.

\paragraph{Build-side observations.} The recursive BFS build's advantage over one-shot construction is similarly stable across LLMs: the HR drop from recursive to one-shot is 32.9 points on V3.2, 32.7 on V4-flash, and narrows to 19.2 on V4-pro. V4-pro's larger capacity does let one-shot design produce a more usable taxonomy (70.5 HR vs.\ V3.2/V4-flash's 59.7--61.6), but it still under-performs the recursive build by a wide margin. The result confirms the Study~2 argument: even a much stronger LLM cannot replace the explicit BFS iteration with a single design call. A second observation cuts deeper on V4: the one-shot A2X taxonomy under-performs even the full-context baseline on every V4 model (e.g., V4-flash 61.6 vs.\ 94.0 HR), so on capable models the \emph{recursive} build is the only configuration in which the taxonomy approach beats simply dumping the registry into the prompt.

\section{Baseline Implementation Details}
\label{sec:appendix_baselines}

We describe each baseline in enough detail to reproduce the numbers in Tables~1--3.

\subsection{Pure-LLM full-context retrieval (Study~1)}
\label{sec:appendix_purellm}

This baseline concatenates every service in the registry (\texttt{id}, \texttt{name}, \texttt{description}) into a single user prompt, prepended by a system instruction asking the LLM to return the IDs of all services relevant to the query. The model is the same DeepSeek-V3 chat endpoint used by A2X. Each query consumes about $67$k input tokens on ToolRet; we evaluate on the full 1{,}714-query test set, with parallelism set to 20 workers (DeepSeek's prefix caching of the shared 1{,}839-service catalog keeps wall-clock cost manageable). The full-1{,}714 result is reported in Table~\ref{tab:study1}; per-method Precision in Table~\ref{tab:appendix_prec_e1}.

\begin{table}[t]
\centering
\small
\begin{tabular}{lcc}
\toprule
\textbf{Metric} & \textbf{Pure-LLM} & \textbf{A2X (\texttt{get\_all})} \\
\midrule
Precision (\%) & 16.8 & 16.2 \\
\bottomrule
\end{tabular}
\caption{Per-method Precision for Table~\ref{tab:study1} (Study~1).}
\label{tab:appendix_prec_e1}
\end{table}

\subsection{One-shot LLM build (Study~2)}
\label{sec:appendix_oneshot}

The one-shot baseline mirrors A2X's output shape (a 3-level hierarchical taxonomy plus per-service assignments) but produces the structure in a \emph{single} LLM call. Specifically:

\begin{enumerate}
\itemsep0pt
\item \textbf{Design (one call).} A single chat-completion request is issued with all 1{,}839 service \texttt{name: description} pairs in the user message. The prompt asks the LLM to emit a strict JSON tree with 6--10 top-level categories, 3--5 subcategories each, and 3--5 leaf categories each. \emph{No} single-axis constraint is supplied; \emph{no} keyword-first compression; \emph{no} refinement loop.
\item \textbf{Classification (one call per service, in parallel).} The taxonomy from step~1 is serialized as an indented bullet list. For each service, a separate LLM call asks for the leaf path (``Top \textgreater\ Sub \textgreater\ Leaf'') the service belongs to. Returned paths are parsed and reconciled against the taxonomy node names; classifications whose path cannot be resolved are recorded as failures.
\item \textbf{Persistence.} The resulting taxonomy plus service assignments are written in A2X's \texttt{taxonomy.json} / \texttt{class.json} format so the same evaluator can score downstream search.
\end{enumerate}

On ToolRet the one-shot pipeline issues 1{,}840 LLM calls (1 design + 1{,}839 classifications) for a total of $\sim$5.5M tokens. The resulting taxonomy has 160 leaves and successfully assigns 1{,}715 of 1{,}839 services; the remaining 124 services fail classification because the unrefined taxonomy lacks a leaf they cleanly map into. Downstream A2X search on the produced taxonomy uses identical search-side code (\texttt{get\_all} mode) as the main A2X result, so the search-quality difference is attributable to the taxonomy itself rather than to the search procedure.

\paragraph{Additive variants.} We additionally evaluate three variants that add a single A2X build module on top of the one-shot baseline. All three reuse steps~2 (parallel classification) and~3 (persistence) above unchanged; only step~1 (design) is modified.
\begin{itemize}
\itemsep0pt
\item \textbf{+ Freq table.} Before the design call, we extract keyword tokens from all service descriptions via A2X's keyword extractor (\texttt{keyword\_threshold}$=$500, \texttt{keyword\_batch\_size}$=$50). The design prompt receives this compressed keyword vocabulary instead of raw descriptions, dropping total cost to $\sim$2.2\,M tokens.
\item \textbf{+ Refine loop.} After the initial design, we run up to \texttt{max\_refine\_iterations}$=$3 refinement cycles. Each cycle re-classifies all 1{,}839 services, scans for misclassified or hard-to-place items, and feeds the diagnostics back as a refinement instruction to the LLM. Total cost rises to $\sim$12.2\,M tokens.
\item \textbf{+ Single-axis.} We prepend the same single-axis rules used by A2X's \texttt{CATEGORY\_DESIGN\_TEMPLATE} (``USER FUNCTIONAL DOMAIN ONLY'') to the one-shot design prompt, forbidding splits along non-functional axes (modality, provider, deployment form, etc.). Total cost: $\sim$6.7\,M tokens.
\end{itemize}

\begin{table}[t]
\centering
\small
\begin{tabular}{lrc}
\toprule
\textbf{Build method}     & \textbf{Build tokens} & \textbf{Precision (\%)} \\
\midrule
One-shot (baseline)       & 5.5\,M                & 12.0 \\
\;+ Freq table            & 2.2\,M                & 20.9 \\
\;+ Refine loop           & 12.2\,M               & 23.2 \\
\;+ Single-axis           & 6.7\,M                & 20.6 \\
\textbf{A2X recursive BFS}& \textbf{29\,M}        & 16.2 \\
\bottomrule
\end{tabular}
\caption{Per-build token cost and Precision for Table~\ref{tab:study2} (Study~2 additive build ablation). A2X's build is a one-time amortized cost.}
\label{tab:appendix_prec_e2}
\end{table}

\subsection{Dense embedding baselines (Study~3)}
\label{sec:appendix_embed}

\begin{table*}[!t]
\centering
\small
\setlength{\tabcolsep}{4pt}
\begin{tabular}{lllrrrrrr}
\toprule
\textbf{Dataset} & \textbf{Method} & \textbf{K} & \multicolumn{3}{c}{\textbf{EN}} & \multicolumn{3}{c}{\textbf{CN}} \\
\cmidrule(lr){4-6} \cmidrule(lr){7-9}
                 &                 &            & HR $\uparrow$ (\%) & Recall $\uparrow$ (\%) & Prec. (\%) & HR $\uparrow$ (\%) & Recall $\uparrow$ (\%) & Prec. (\%) \\
\midrule
\multirow{11}{*}{ToolRet}   & \multirow{2}{*}{MiniLM}                & 5  & 69.1 & 61.8 & 15.2 & --   & --   & --   \\
                            &                                        & 10 & 76.0 & 69.3 &  8.7 & --   & --   & --   \\
                            & \multirow{2}{*}{BGE-large-en}          & 5  & 68.0 & 61.7 & 15.3 & --   & --   & --   \\
                            &                                        & 10 & 75.5 & 69.2 &  8.7 & --   & --   & --   \\
                            & \multirow{2}{*}{text2vec-base-chinese} & 5  & --   & --   & --   & 41.4 & 36.9 &  8.9 \\
                            &                                        & 10 & --   & --   & --   & 49.6 & 44.3 &  5.4 \\
                            & \multirow{2}{*}{BGE-M3}                & 5  & 64.8 & 58.0 & 14.5 & 60.9 & 54.4 & 13.5 \\
                            &                                        & 10 & 72.3 & 65.9 &  8.4 & 71.4 & 64.9 &  8.2 \\
                            & \multirow{2}{*}{MCP-Zero}              & 5  & 62.1 & 55.9 & 13.6 & 58.6 & 52.4 & 12.8 \\
                            &                                        & 10 & 68.8 & 62.4 &  7.8 & 62.2 & 56.0 &  6.9 \\
                            & \textbf{A2X}                           & -- & \textbf{92.6} & \textbf{89.2} & 16.2 & \textbf{90.3} & \textbf{86.4} & 12.0 \\
\midrule
\multirow{11}{*}{publicMCP} & \multirow{2}{*}{MiniLM}                & 5  & 72.0 & 41.1 & 21.2 & --   & --   & --   \\
                            &                                        & 10 & 80.0 & 49.8 & 13.0 & --   & --   & --   \\
                            & \multirow{2}{*}{BGE-large-en}          & 5  & 60.0 & 32.7 & 17.6 & --   & --   & --   \\
                            &                                        & 10 & 70.0 & 40.5 & 11.0 & --   & --   & --   \\
                            & \multirow{2}{*}{text2vec-base-chinese} & 5  & --   & --   & --   & 20.0 &  7.6 &  4.4 \\
                            &                                        & 10 & --   & --   & --   & 28.0 & 11.4 &  3.2 \\
                            & \multirow{2}{*}{BGE-M3}                & 5  & 66.0 & 35.4 & 18.8 & 58.0 & 29.1 & 16.0 \\
                            &                                        & 10 & 70.0 & 41.8 & 11.4 & 74.0 & 43.3 & 11.8 \\
                            & \multirow{2}{*}{MCP-Zero}              & 5  & 54.0 & 24.7 & 14.4 & 40.0 & 18.7 & 10.8 \\
                            &                                        & 10 & 52.0 & 27.5 &  7.6 & 48.0 & 23.3 &  7.0 \\
                            & \textbf{A2X}                           & -- & \textbf{100.0} & \textbf{94.9} & 10.5 & \textbf{98.0} & \textbf{88.9} &  9.9 \\
\bottomrule
\end{tabular}
\caption{\textbf{Full top-$K$ comparison at $K{=}5$ and $K{=}10$ on ToolRet and publicMCP.}}
\label{tab:appendix_topk}
\end{table*}

All dense baselines share the same pipeline: encode every service description with the chosen embedding model, build a flat index, then for each query encode the raw query text and return the top-$K$ cosine-nearest services; per-dataset $K$ values are as reported in Table~\ref{tab:study3} caption, and the full $K$ sweep is in Table~\ref{tab:appendix_topk}. We tested four embedding models:
\texttt{sentence-transformers/all-MiniLM-L6-v2} (384d, English),
\texttt{BAAI/bge-large-en-v1.5} (1{,}024d, English),
\texttt{shibing624/text2vec-base-chinese} (768d, Chinese),
and \texttt{BAAI/bge-m3} (1{,}024d, multilingual).
Service and query embeddings are cached on disk so repeated runs at different $K$ values do not re-encode; all vectors are L2-normalized explicitly before cosine similarity. Similarity computation uses NumPy dense matrix multiplication.

\begin{table}[t]
\centering
\small
\begin{tabular}{lcc}
\toprule
\textbf{Method (lang.)}        & \textbf{ToolRet} & \textbf{publicMCP} \\
\midrule
MiniLM (EN)                    & 15.2             & 21.2 \\
BGE-large-en (EN)              & 15.3             & 17.6 \\
text2vec-base-chinese (CN)     &  8.9             &  4.4 \\
BGE-M3 (EN)                    & 14.5             & 18.8 \\
BGE-M3 (CN)                    & 13.5             & 16.0 \\
MCP-Zero (EN)                  & 13.6             & 14.4 \\
MCP-Zero (CN)                  & 12.8             & 10.8 \\
\textbf{A2X (EN)}              & \textbf{16.2}    & \textbf{10.5} \\
\textbf{A2X (CN)}              & \textbf{12.0}    & \textbf{ 9.9} \\
\bottomrule
\end{tabular}
\caption{Per-method Precision (\%) for Table~\ref{tab:study3} (Study~3 embedding-based retrieval), at the per-dataset $K$ used in Table~\ref{tab:study3} ($K{=}5$ on ToolRet, $K{=}10$ on publicMCP).}
\label{tab:appendix_prec_e3}
\end{table}

Table~\ref{tab:appendix_topk} reports the full top-$K$ comparison at both $K{=}5$ and $K{=}10$ on both datasets; the rows that feed into Table~\ref{tab:study3} are $K{=}5$ on ToolRet and $K{=}10$ on publicMCP.

\subsection{MCP-Zero baseline (Study~3)}
\label{sec:appendix_mcpzero}

We follow the reference implementation released by \citet{fei2025mcpzero} (\url{https://github.com/xfey/MCP-Zero}). For each query we issue an LLM call with the \emph{system\_ours\_mcptools.prompt} system prompt and the \emph{user\_query\_without\_server.prompt} user template (\textit{``I need to \{task\_description\}. Please identify the most appropriate tool for this task.''}). The LLM is instructed to reply with a block of the form
\begin{quote}
\small
\texttt{<tool\_assistant>}\\
\texttt{server: [platform / service domain]}\\
\texttt{tool: [specific tool the user needs]}\\
\texttt{</tool\_assistant>}
\end{quote}
We parse the response, extract the \texttt{tool: ...} description, embed it with \texttt{BAAI/bge-m3}, and rank services by cosine similarity to the precomputed service embeddings, returning the top-$K$ under the same per-dataset $K$ policy as \S\ref{sec:appendix_embed}. We use MCP-Zero's flat (single-stage) retrieval configuration (the \texttt{experiment\_apibank.py} script in the released code), as ToolRet and publicMCP have flat service inventories with no server / tool hierarchy. The system prompt, user template, and \texttt{<tool\_assistant>} parsing logic follow the released code.